\def\year{2020}\relax
\documentclass[letterpaper]{article} 
\usepackage{aaai20}  
\usepackage{times}  
\usepackage{helvet} 
\usepackage{courier}  
\usepackage[hyphens]{url}  
\usepackage{graphicx} 
\usepackage{graphicx}  
\usepackage{amssymb}
\usepackage{amsmath}
\usepackage{tabularx}
\usepackage{multirow}
\usepackage{booktabs}
\usepackage{amsfonts}
\usepackage{footnote}
\title{Unicoder-VL: A Universal Encoder for Vision and Language by Cross-modal Pre-training}
\author{
Gen Li,\textsuperscript{\rm 1}\thanks{Work is done during an internship at Microsoft Research Asia.} 
Nan Duan,\textsuperscript{\rm 2}
Yuejian Fang,\textsuperscript{\rm 1}\thanks{Corresponding author.} Ming Gong,\textsuperscript{\rm 3} Daxin Jiang,\textsuperscript{\rm 3} Ming Zhou\textsuperscript{\rm 2}  \\
\textsuperscript{\rm 1}School of Software \& Microelectronics, Peking University, Beijing, China  \\
\textsuperscript{\rm 2}Natural Language Computing, Microsoft Research Asia, Beijing, China \\
\textsuperscript{\rm 3}STCA NLP Group, Microsoft, Beijing, China \\
ligen.li@pku.edu.cn, fangyj@ss.pku.edu.cn \\
\{nanduan, migon, djiang, mingzhou\}@microsoft.com 
}
 
\begin{document}

\maketitle

\begin{abstract}
We propose \textbf{Unicoder-VL}, a universal encoder that aims to learn joint representations of vision and language in a pre-training manner. 
Borrow ideas from cross-lingual pre-trained models, such as XLM \cite{lample2019cross} and Unicoder \cite{huang2019unicoder}, both visual and linguistic contents are fed into a multi-layer Transformer \cite{vaswani2017attention} for the cross-modal pre-training, where three pre-trained tasks are employed, including Masked Language Modeling (MLM), Masked Object Classification (MOC) and Visual-linguistic Matching (VLM).
The first two tasks learn context-aware representations for input tokens based on linguistic and visual contents jointly. The last task tries to predict whether an image and a text describe each other. 
After pretraining on large-scale image-caption pairs, we transfer Unicoder-VL to caption-based image-text retrieval and visual commonsense reasoning, with just one additional output layer. We achieve state-of-the-art or comparable results on both two tasks and show the powerful ability of the cross-modal pre-training.
\end{abstract}
\section{Introduction}

In recent years, pre-trained models have made great progress in both computer vision (CV) and natural language processing (NLP) communities.

In CV, pre-trained models, such as VGG \cite{simonyan2014very} and ResNet \cite{he2016deep}, are usually trained based on CNN using ImageNet \cite{deng2009imagenet}, whose training objective is to predict the categorical label of a given image. 
For downstream tasks, such as image classification, image retrieval \cite{karpathy2015deep} \cite{lee2018stacked} and object detection \cite{ren2015faster}, the resulting models can extract feature representations for input images, which will be further used in following task-specific models. 

In NLP, pre-trained models, such as BERT \cite{devlin2018bert}, XLNet \cite{yang2019xlnet} and RoBERTa \cite{liu2019roberta}, have achieved state-of-the-art performances in many NLP tasks as well, such as natural language inference \cite{bowman2015large}, and machine reading comprehension \cite{rajpurkar2016squad}. 
Pre-trained with language modeling, such models can learn general knowledge from large-scale corpus first, and then transfer them to downstream tasks with simple fine-tuning layers.

However, these two types of pre-trained models cannot well handle a cross-modal task directly, if its natural language inputs are long sequences (such as questions), rather than short phrases (such as tags). The reason is two-fold. On one hand, as ImageNet covers categorical labels only, the resulting models cannot deal with long sequences. This is why most such tasks, e.g. visual question answering (VQA) \cite{antol2015vqa}, visual commonsense reasoning (VCR) \cite{zellers2019recognition} and image retrieval \cite{karpathy2015deep}, still need additional fusion layers to model interaction between visual and linguistic contents.
On the other hand, existing NLP pre-trained models can handle long natural language sequences very well. But none of them is trained with visual contents directly.

\begin{figure*}[!htbp]
	\centering
	\includegraphics[width=\textwidth]{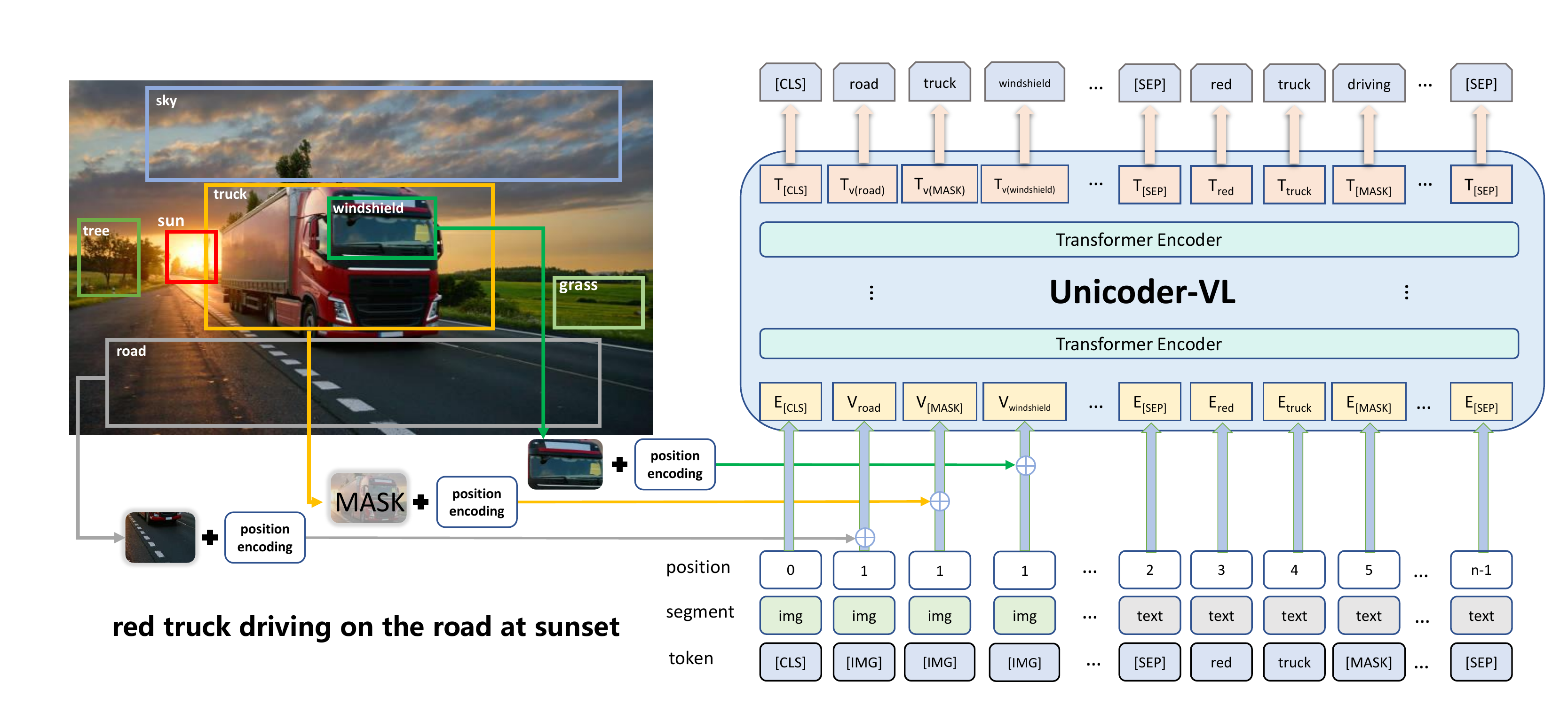}
	\caption{Illustration of Unicoder-VL in the context of an object and text masked token prediction, or $cloze$, task. Unicoder-VL contains multiple Transformer encoders which are used to learn viusal and linguistic representation jointly.}
	\label{fig:pretraintask}
\end{figure*}

Motivated by these, we propose a \textbf{Uni}versal en\textbf{coder} for \textbf{V}ision and \textbf{L}anguage, short for \textbf{Unicoder-VL}, a universal encoder based on a multi-layer Transformer \cite{vaswani2017attention}, which aims to learn joint representations of vision and language (especially for long sequences) in a pre-training manner.
Inspired by BERT and some recent cross-lingual pre-trained models. such as XLM \cite{lample2019cross} and Unicoder (Huang et al., 2019), a cross-modal pre-training framework is designed to model the relationships between visual and linguistic contents and learn their joint representations.
We use large-scale image-caption pairs in Unicoder-VL training, as such annotations are easy to collect from web, with relatively good quality. 
Three pre-trained tasks are employed, including Masked Language Modeling (MLM), Masked Object Classification (MOC) and Visual-linguistic Matching (VLM). The first two tasks learn context-aware representations for input tokens based on linguistic and visual contents jointly. The last task tries to predict whether an image and a text describe each other.

As the first step along this new pre-training direction, we evaluate Unicoder-VL on image-text retrieval tasks. From experiments we can see that, by adding a simple fine-tuning layer, Unicoder-VL achieves state-of-the-art results on both MSCOCO \cite{chen2015microsoft} and Flicker30K \cite{young2014image}, comparing to a bunch of strong baselines. Furthermore, it also shows good performance in a zero-shot setting, which indicates a generalization ability. In VCR, we achieve comparable results with concurrent state-of-the-art works. It shows that cross-modal pre-training improve the ability of visual commonsense reasoning.

The main contributions of our work are summarized as follows. We leverage a multi-layer Transformer to model cross-modal semantic representations. Meanwhile, we propose three well-designed cross-modal pre-training tasks to learn high-level visual representations and capture rich relationships between visual and linguistic contents. We fine-tune our pre-trained model to image-text retrieval and visual commonsense reasoning task and achieve significant improvements, demonstrating the effectiveness of our proposed method. 
Note, this pre-training method is general and not limited to image-text retrieval tasks. We will move further to evaluate it on more cross-modal tasks, such as image captioning \cite{anderson2018bottom}, scene graph generation, video classification and video question answering.

\section{Related Work}

\subsection{Pre-training for CV Tasks}
Most existing pre-trained CV models are based on multi-layer CNN, such as VGG (Simonyan and
Zisserman 2014) and ResNet (He et al. 2016), and trained using ImageNet. As ImageNet \cite{deng2009imagenet} only contains image labels, the resulting pre-trained models cannot deal with cross-modal tasks with long natural language inputs, such as queries in image retrieval and VQA tasks. These tasks pay more atttention on visual relations and descriptions rather than what is the image.
By contrast, Unicoder-VL is pre-trained using image-caption pairs. So it is more suitable to these tasks. 
\subsection{Pre-training for NLP Tasks}
Latest pre-trained NLP models are based on multi-layer Transformer, such as GPT \cite{radford2018improving}, BERT \cite{devlin2018bert}, XLNet (Yang et al., 2019) and RoBERTa\cite{liu2019roberta}. All of the works are trained using large-scale corpus by language modeling. Such models learn contextualized text representations by predicting word tokens based on their contexts, and can be adapted to downstream tasks by additional fine-tuning. 

Since the image is not a sequential data, the autoencoding objective of BERT is very appropriate for visual content. The key question is how to include visual contents in pre-training as well. However, the cross-modal pre-training is not limited to transformer-based models like BERT or XLNet. We leave more exploration in the future.

\subsection{Pre-training for Cross-modal Tasks}
Very recently, several attempts have been made to pre-train models for cross-modal tasks. 

VideoBERT \cite{sun2019videobert} is one such method, whose goal is to learn cross-modal representations from videos and their corresponding transcripts. However, instead of using visual features directly in pre-training, it generates a sequence of “visual words” from each video first, and then uses them with transcript words together in LM pre-training. While in Unicoder-VL, we present visual features of objects in the images jointly training with linguistic contents.

Concurrent to our work, several recent released works, such as ViLBERT \cite{lu2019vilbert}, VisualBERT \cite{li2019visualbert}, VL-BERT \cite{su2019vl} and UNITER \cite{chen2019uniter} are pre-training methods on vision-and-language tasks. The concurrent emergency of these research works indicates the importance of deriving a generic pre-trainable representation for cross-modal tasks.

The comparison of these models are: 

1) The model of ViLBERT is a two single-modal network applied on input sentences and images respectively, followed by a cross-modal Transformer combining information from the two sources. They propose a co-attentional Transformer layer (Co-TRM) in their model and claim such structure has a better ability to model interactions between visual and linguistic contents. Then the third Transformer fuses them. On the other hand, VisualBERT, Unicoder-VL, VL-BERT and UNITER proposed a single-stream architecture (vanilla BERT structure), which fuses cross-modal information early and freely,

2) a) The masked language model pre-training task is used by all of the above models. b) VisualBERT does not apply the object prediction task. Our model predicts the object labels while the others calculate the KL divergence between the input and output distributions. VL-BERT masked the image before applying by Faster-RCNN. UNITER masks only one modality each time. c) The visual-linguistic matching is used by all of the above models except VL-BERT, which claims this task is of no use.

3) VisualBERT is pre-trained on MSCOCO Captions dataset. ViLBERT, and VL-BERT are all pre-trained on about 3 million Conceptual Captions \cite{sharma2018conceptual} dataset and then transfer to down-stream tasks. UNITER add about 1 million image-caption pair besides the Conceptual Captions and in-domain  MSCOCO Caption and Visual Genome Dense Captions \cite{krishna2017visual} data.

Compare to recent works, we achieve the SOTA results on image-to-text and text-to-image retrieval and VCR, which proves Unicoder-VL's ability on these tasks.

\section{Approach}

\noindent In this section, we first briefly summarize the original BERT model, and present our cross-modal pre-trained model Unicoder-VL, including details of image and text pre-processing and three cross-modal pre-training tasks we used. 

\subsection{BERT}
BERT \cite{devlin2018bert} is a pre-trained model based on multi-layer Transformer \cite{vaswani2017attention}. Two tasks are used in pre-training: masked language model and next sentence prediction. 
In masked language model, BERT tries to predict the identity of each masked word based on all context words. 
In next sentence prediction, BERT tries to predict whether the second half of the input follows the first half of the input in the corpus, or is a random paragraph. A special token, \texttt{[CLS]}, is prepended to every input sequence, and its representation in final layer will be used for the next sentence prediction task.

\subsection{Unicoder-VL}
The overview of Unicoder-VL is shown in Fig~\ref{fig:pretraintask}. Given a pair of image and sentence, Unicoder-VL takes the visual regions of the image and textual tokens of the sentence as the input and then encode the input to the  linguistic embedding and image embedding. These embeddings are then fed into a multi-layer self-attention Transformer to learn a cross-modality contextualized embedding between visual regions and textual tokens.

\textbf{Linguistic Embedding.} Following the text pre-processing of BERT, We tokenize each input text \textbf{w} $=\{w_1,...,w_T\}$. $T$ is the length of the WordPiece \cite{wu2016google} linguistic input. Besides, as shown in  Fig~\ref{fig:pretraintask}, we also add the special tokens \texttt{[CLS]} and \texttt{[SEP]}. For the visual elements, a special \texttt{[IMG]} token is assigned for each one of them. The final representation for each sub-word token is obtained via summing up its word embedding and position embedding, followed by a layer normalization (LN) layer. These embeddings are all initialized from BERT.

\textbf{Image Embedding.} For each input image, we first use Faster R-CNN (weights are initialized from \cite{singh2018pythia}) to extract the visual features (pooled ROI features) for each region. We also encode the location features with a 5-D vector, $b=(\frac{x_1}{W}, \frac{y_1}{H}, \frac{x_{2}}{W}, \frac{y_{2}}{H}, \frac{(y_{2}-y_{1})(x_{2}-x_{1})}{W \cdot H})$, where $(x_1, y_1)$ and $(x_2, y_2)$ denote the coordinate of the bottom-left and top-right corner and the fraction of image area covered respectively, and $W$, $H$ are of the width and height of the input image. Both visual and location features are then fed through a fully-connected (FC) layer, to be projected into the same embedding space. The final visual embedding for each region is obtained by summing up the two FC outputs and then passing through another LN layer. The final image regions are denotes as \textbf{v} $=\{v_1,...,v_I\}$. $I$ is the length of the objects extracted from this image.

We also keep the predicted label of each detected object, which will be used in the object label prediction task. Note that the whole Faster R-CNN model is fixed during training. 

\textbf{Pre-training Tasks.} We propose three tasks when doing the cross-modal pre-training: Masked Language Modeling (MLM), Masked Object Classifation (MOC) and Visual-linguistic Matching (VLM). 

\textbf{Masked Language Modeling (MLM)}. We denote the linguistic input as \textbf{w} $=\{w_1,...,w_T\}$ and object regions as \textbf{v} $=\{v_1,...,v_I\}$, and the mask indices as $\textbf{\rm{m}} \in \mathbb{N}^{M}$. In MLM, we randomly mask out the input words with probability of 15\%, and replace the masked ones $\rm{\textbf{w}}_{\rm{\textbf{m}}}$ with special token \texttt{[MASK]}. The goal is to predict these masked words based on the observation of their surrounding words $\rm{\textbf{w}}_{\backslash \rm{\textbf{m}} }$ and all image regions \textbf{v}, by minimizing the negative log-likelihood:
\begin{equation}
\mathcal{L}_{ \rm{MLM}}(\theta)=-E_{(\rm{\textbf{w}}, \rm{\textbf{v}})\sim D} \log P_{\theta}(\rm{\textbf{w}}_{\rm{\textbf{m}}}|\rm{\textbf{w}}_{\backslash \rm{\textbf{m}} }, \textbf{v})
\end{equation}
where $\theta$ is the trainable parameters. Each pair (\textbf{w}, \textbf{v}) is sampled from the whole training set $D$.

\textbf{Masked Object Classifation (MOC)}. Similar to MLM, we also sample image regions and mask their visual features with a probability of 15\%. We replace the object feature vector with a zero-initialized vector $\rm{\textbf{v}}_{\rm{\textbf{m}}}$ 90\% of the time, and keep the object feature unchanged in the left 10\% time. We simply take the object category with the highest confidence score predicted by the same detection model as the ground-truth label. We first feed the Transformer output of the masked region $\rm{\textbf{v}}_{\rm{\textbf{m}}}^{(i)}$
m into an FC layer
to predict the scores of K object classes, which further goes through a softmax function to be transformed into a normalized distribution $g_{\theta}(\rm{\textbf{v}}_{\rm{\textbf{m}}}^{(i)})$.  The final objective is:
\begin{equation}
\mathcal{L}_{ \rm{MOC}}(\theta)=-E_{(\rm{\textbf{w}}, \rm{\textbf{v}})\sim D} \sum_{i=1}^{M}{\rm{CE} (c(\rm{\textbf{v}}_{\rm{\textbf{m}}}^{(i)}), g_{\theta}(\rm{\textbf{v}}_{\rm{\textbf{m}}}^{(i)}))}
\end{equation}
where $c(\rm{\textbf{v}}_{\rm{\textbf{m}}}^{(i)}) \in \mathbb{R}^K$ is the one-hot vector of the ground-truth label.

\textbf{Visual-linguistic Matching (VLM)}. we also learn an instance-level alignment (rather than token/region-level) between the whole image and the sentence via VLM. We take final hidden state of \texttt{[CLS]} to predict whether the linguistic sentence is semantically matched with the visual content, with an additional FC layer. The scoring function is denoted as $s_{\theta}(\rm{\textbf{w}},\rm{\textbf{v}})$. During training, we sample both positive and negative image-sentence pairs and learn their matching scores (including negative image and negative sentence). We denote the label as $y \in \{0, 1\}$, indicating if the sampled pair is a match. Then

\begin{equation}
\begin{split}
\mathcal{L}_{ {\rm VLM}}(\theta)=&-E_{({\rm\textbf{w}}, {\rm\textbf{v}})\sim D}[y\log s_{\theta}({\rm\textbf{w}}, {\rm\textbf{v}}) \\
& + (1-y) \log (1-s_{\theta}(\rm{\textbf{w}}, \rm{\textbf{v}}))]
\end{split}
\end{equation}


Overall, we have three training regimes corresponding to the image-text inputs. Our final pre-training objective is the sum of the losses above:
\begin{equation}
\mathcal{L}=(\mathcal{L}_{{\rm MLM}} + \mathcal{L}_{{\rm MOC}})\cdot I[y=1] + \mathcal{L}_{{\rm VLM}}
\end{equation}
where $I[y=1]$ is an indicator for the label 1 being correct for the image and caption pair.

\section{Experiments}

\noindent In this section, we describe how we pre-train our model and show the evaluation details on image-text retrieval task to which we transfer the pre-trained model. 
\subsection{Pre-training Unicoder-VL}

Conceptual Captions dataset \cite{sharma2018conceptual} contains about 3.3M image and caption pairs harvested from the web, which are very suitable for our cross-modal pre-training. Due to some broken urls, the size of image-caption pairs of Conceptual Captions dataset is about 3M.

Similar to Conceptual Captions, SBU Captions  \cite{ordonez2011im2text} dataset is also automatically collected from Web and contains 1M image-caption pairs. Due to some broken urls, the size of image-caption pairs of SBU dataset is about 0.8M. 

Finally, we use 3.8M image-caption pairs to do pre-training.

Our model has 12 layers of Transformer blocks, where each block has 768 hidden units and 12 self-attention heads. The maximum sequence length is set as 144. We sample 1 negative image or 1 negative caption and then judge whether this image and caption is matching when do the VLM task. The parameters are initialized from BERT-base, which is pre-trained on text data only. 

For the visual part, we use fixed 100 RoIs with detection scores higher
than 0.2 are selected for each image. If eligible RoIs are less than 100, we simply select the top-100 RoIs, regardless of the detection score threshold.

During Pre-training, our experiments are running on 4 NVIDIA Tesla V100 GPU. Our best performing model is pre-trained for 10 epochs with three training tasks introduced above, using the ADAM optimizer with learning rate of 1e-4 with a batch size of 192 with gradient accumulation (every 4 steps). The model will warmup the first 10\% of all training steps. We use float16 operations to speed up training and to reduce the memory usage of our models. 

\begin{table*}[!htbp]
\resizebox{0.98\textwidth}{!}{
\begin{tabular}{lcccccccccccc}
 \toprule
\multirow{3}{*}{Methods}
& \multicolumn{6}{c}{MSCOCO} 
& \multicolumn{6}{c}{Flickr30k} \\
\cmidrule(lr){2-7} \cmidrule(l){8-13}
& \multicolumn{3}{c}{Sentence Retrieval} 
& \multicolumn{3}{c}{Image Retrieval} 
& \multicolumn{3}{c}{Sentence Retrieval} 
& \multicolumn{3}{c}{Image Retrieval} \\
\cmidrule(lr){2-4} \cmidrule(l){5-7} \cmidrule(lr){8-10} \cmidrule(l){11-13} 
 &  R@1 & R@5 & R@10 & R@1 & R@5 & R@10 &  R@1 & R@5 & R@10 & R@1 & R@5 & R@10\\
\midrule 
\multicolumn{13}{c}{1K Test set } \\

  \midrule
   DVSA~\cite{karpathy2015deep}  &	38.4 &	69.9 &	80.5  &	27.4 &	60.2 &	74.8  &	22.2 &48.2 &61.4  &	15.2 &	37.7 &	50.5 \\
    m-CNN~\cite{ma2015multimodal}  &	42.8 &	73.1 &	84.1  &	32.6 &	68.6 &	82.8 &33.6 &	64.1 &74.9  &26.2 &	56.3 &69.6    \\
    DSPE~\cite{wang2016learning}  &	50.1 &	79.7 &	89.2  &	39.6 &	75.2 &	86.9 &40.3 &68.9 &79.9  &29.7 &60.1 &72.1  \\
    VSE++~\cite{faghri2017vse++} &	64.7 &	- &	95.9  &	52.0 & - &	92.0  &52.9 &79.1 &87.2  &39.6 &69.6 &79.5 \\
   SCAN~\cite{lee2018stacked} &	72.7 &	94.8 &	98.4  &	58.8 &	88.4 &	94.8 &67.4 &90.3 &95.8  &48.6 &	77.7 &85.2 \\
   SCG~\cite{ijcai2019-720}  &	76.6 &	96.3 &	99.2  &	61.4 &	88.9 &	95.1 &71.8 &90.8 &94.8 &49.3 &76.4 &85.6 \\
    PFAN~\cite{ijcai2019-526}  & 76.5 &	96.3 &	99.0  &	61.6 &	89.6 &	95.2 &70.0 &91.8 &95.0 &50.4 &78.7 &86.1 \\
ViLBERT~\cite{lu2019vilbert}$^\dag$ &- &- &- &- &- &- &- &- &- &58.2 &84.9 &91.5 \\
UNITER~\cite{chen2019uniter}$^\dag$ &- &- &- &- &- &- &84.7 &\textbf{97.1} &99.0 &71.5 &\textbf{91.2} &\textbf{95.2} \\
  \midrule
    Unicoder-VL (zero-shot)  &54.4 &82.8 & 90.6  &43.4 &76.0 &87.0 &64.3 &85.8 & 92.3  &48.4 &76.0 &85.2 \\
    Unicoder-VL (w/o pre-training)  &	75.1 &94.3 & 97.8  &63.9 &91.6 &96.5 &73.0 &89.0 & 94.1  &57.8 &82.2 &88.9 \\
    Unicoder-VL  &	\textbf{84.3} &\textbf{97.3} &\textbf{99.3}  &	\textbf{69.7} &\textbf{93.5} &\textbf{97.2} &	\textbf{86.2} &96.3 &\textbf{99.0}  &\textbf{71.5} &90.9 &94.9 \\
    \midrule 
\multicolumn{13}{c}{5K Test set } \\
  \midrule
SCAN~\cite{lee2018stacked}  & 50.4 &	82.2 &	90.0  & 38.6 &	69.3 &	80.4 &- &- &- &- &- &-  \\
  SCG~\cite{ijcai2019-720}  &56.6 &84.5 &92.0  &39.2 &	68.0 &	81.3 &- &- &- &- &- &-  \\
UNITER~\cite{chen2019uniter}$^\dag$  &\textbf{63.3}  &87.0 &\textbf{93.1}  &\textbf{48.4} &\textbf{76.7} &	\textbf{85.9} &- &- &- &- &- &-  \\
  \midrule
Unicoder-VL  &62.3 &\textbf{87.1} & 92.8  &46.7 &76.0 &85.3 &- &- &-  &- &- &- \\
  \bottomrule 
 \end{tabular}}
\caption{Evaluation results on MSCOCO and Flickr30k test set.$^\dag$ means the concurrent work.}
\label{retrievalresult}
\end{table*}

\subsection{Fine-tune on Downstream Tasks}
The pre-trained Unicoder-VL model can be transferred to multiple downstream visual-linguistic tasks,
with simple modifications on the input format, output prediction, loss function and training strategy.
\subsubsection{Image-Text Retrieval.}
Image-text retrieval is the task of identifying an image from candidates given a caption describing its content, or vice versa. We use two datasets as follows. 1) MSCOCO consists of 123,287 images, and each image contains roughly five textual descriptions. It is split into 82,783 training images, 5,000 validation images and 5,000 testing images. We follow the data split in \cite{faghri2017vse++} to add 30,504 images that were originally in the validation set of MSCOCO. 2) Flickr30K contains 31,783 images collected from the Flickr website. Following \cite{karpathy2015deep}, we split the dataset into 29,783 training images, 1,000 validation images and 1,000 testing images. Besides, we use three evaluation metrics, i.e., R@K (K=1,5,10). R@K is the percentage of ground-truth matchings appearing in the top K-ranked results. 

During fine-tuning on image-text retrieval, we formulate it as a ranking problem. we sample 3 negative cases in each matching tasks. 
Inputs of fine-tuning share the same data preprocessing procedures with pre-training, except that we do not mask word and object in the fine-tuning stage. Similar to the VLM task, we also denote the score function as $s_{\theta}(\rm{\textbf{w}},\rm{\textbf{v}})$. We omit this trainable parameter $\theta$ below. We propose two image-text matching tasks: image-to-text, text-to-image. We use triplet loss and maximize the margin of positive and negative samples after generating the similarity score between two input modalities. 

In this study, we focus on the hardest negatives in every sampled examples, following \cite{faghri2017vse++}. For a positive pair $({\rm \textbf{w}},{\rm \textbf{v}})$, the hardest negatives are given by ${\rm \textbf{v}}_h^-=\arg\max_{{\rm \textbf{v}}_i\neq {\rm \textbf{v}}} s({\rm \textbf{w}}, {\rm \textbf{v}}_i)$ and ${\rm \textbf{w}}_h^-=\arg\max_{{\rm \textbf{w}}_i\neq {\rm \textbf{w}}} s({\rm \textbf{w}}_i, {\rm \textbf{v}})$. So the hardest triplet loss function is:
\begin{equation}
    \mathcal{L}_{hard}(\textbf{x},\textbf{y}) = \sum_{{\rm \textbf{y}}^{-}\in \mathcal{N}_\textbf{y}}{\{\max[0, \gamma-s(\rm{\textbf{x}},\rm{\textbf{y}}) + s(\rm{\textbf{x}},\rm{\textbf{y}_{h}^{-}})]\}}
\end{equation}   
where \textbf{x} and \textbf{y} are encodings of two modality, $\mathcal{N}_{\rm \textbf{y}}$ is the set of negative samples of \textbf{y}. 



Finally, we merge these ranking constraints into one loss function:
\begin{equation}
\mathcal{L} = \lambda_1 \sum_{\textbf{w},\textbf{v}} {\mathcal{L}_{hard}(\textbf{w},\textbf{v})} + \lambda_2 \sum_{\textbf{w},\textbf{v}} {\mathcal{L}_{hard}(\textbf{v},\textbf{w})}
\end{equation}
Followed We use $\gamma = 0.2$, $\lambda_1=1.0$,  $\lambda_2=1.0$ as the hyper-parameters of loss function. The optimizer is Adam and learning rate is set as 5e-5. The batch size is 192 with gradient accumulation (every 4 steps). We also use float16 operations to speed up training and to reduce the memory usage of our models. 

\subsubsection{Zero-shot Image-Text Retrieval.}
The previous tasks are all transferring tasks that include dataset specific fine-tuning. In this zero-shot task, we directly apply the pretrained the multi-modal alignment prediction mechanism to image-text retrieval  without finetuning. The goal of this task is to demonstrate that the pretraining
has developed the ability to ground text and that this can generalize to visual and linguistic variation
without any task specific fine-tuning. We directly use the pre-trained Unicoder-VL model and the same alignment prediction objective as a scoring function
and test on the same split as the image-text retrieval task described above. 

\subsubsection{Visual Commonsense Reasoning.} Given an image, the VCR task presents two problems – visual question answering (Q$\rightarrow$ A) and answer justification (QA$\rightarrow$ R) – both being posed as multiple choice problems. The holistic setting (Q$\rightarrow$AR) requires both the chosen answer and then the
chosen rationale to be correct. The Visual Commonsense Reasoning (VCR) dataset consists of 290k multiple choice QA problems derived from 110k movie scenes. Different from the VQA dataset,
VCR integrates object tags into the language providing direct grounding supervision and explicitly excludes referring expressions. 

To finetune on this VCR, we concatenate the question and each possible response with  semicolons to form four different linguistic inputs and pass each through the model along with the
image. ${\rm \textbf{w}} = \{q_1,...,q_n,;,a_1,...,a_n\}$ for Q$\rightarrow$ A and  ${\rm \textbf{w}} = \{q_1,...,q_n,;,a^{*}_{1},...,a^{*}_{n},;,r_1,...,r_n\}$. Here, $q_0$,... are all question tokens, $a_0$,... are answer tokens, $a^{*}_{0}$ are answer tokens for the correct answer, and $r_0$ are rationale tokens. 

VCR provides ground truth boxes. For each ground truth box, we select the visual feature with highest intersection over union(IoU) from 100 boxes we extract as the new features. Then we add other visual features left after the features with ground truth boxes until the number is 100.

Since some of objects are referenced in $Q$, $A$, $R$, we add visual feature ${\rm \textbf{v}}_i$ to these tokens additionally. $i$ is the object index referenced by the linguistic word.

We also add a projection layer to calculate the score for each pair and the final prediction is a softmax over these four scores. The model is trained under a cross-entropy loss. We trained over 20 epochs with a batch size of 48 and initial learning rate of 3e-5.

\section{Results and Analysis}

\subsection{Evaluation Results}

\subsubsection{Results on Image-Text Retrieval.}
We compare Unicoder-VL with state-of-the-art methods on image retrieval and sentence retrieval tasks in three different settings:
\begin{itemize}
    \item \textbf{zero-shot}, where Unicoder-VL is applied to test set directly, without fine-tuning;
    \item \textbf{task-specific train}, where Unicoder-VL is trained on task-specific training data directly, without pre-training;
    \item \textbf{pre-train + fine-tune}, where Unicoder-VL is further fine-tuned on specific tasks.
\end{itemize}

Experimental results of both datasets are shown in Tab~\ref{retrievalresult}.  
From Tab~\ref{retrievalresult}, the results of the zero-shot setting show that Unicoder-VL can learn general cross-modal knowledge, which take effects in image retrieval and sentence retrieval tasks directly, without any task-specific fine-tuning. Because the difference between automatically collected Conceptual Captions and human-annotated MSCOCO/Flickr30k, this zero-shot result is lower than the finetuned result. Usually finetuning will help the pre-trained model adapt to a little different downstream dataset.

The results of the task-specific train setting show that Unicoder-VL trained on task-specific training data without pre-training still perform better than most previous approaches. It demonstrates the effectiveness of the self-attention mechanism itself on the image-text retrieval tasks.

The results of the pre-train + fine-tune setting show that this setting can significantly outperform all baselines on all evaluation metrics, which proves the superiority of our cross-modal pre-training method. 

Taking R@1 for example, our best result on MSCOCO 1K test set obtains 7.8\% and 8.1\% absolute improvements against the PFAN approach on sentence retrieval task and image retrieval task, respectively. For MSCOCO 5K test set, we can also significantly outperform all baselines on these two tasks. On the Flickr30k testing set, the experiments show similar achievement. Unicoder-VL achieves new state-of-the-art performance and yield a result of 86.2\% and 71.5\% on R@1 for sentence retrieval and image retrieval, respectively. Compared with PFAN, we achieve absolute boost of 16.2\% on R@1 for sentence retrieval and 21.1\% on R@1 for image retrieval. The higher improvement on Flickr30k proves that low-resource task can be improved better with pre-training.

We also compare Unicoder-VL with ViLBERT \cite{lu2019vilbert} and UNITER\cite{chen2019uniter}in the image retrieval and sentence retrieval setting. 10.1 points improvements than ViLBERT show the superiority of Unicoder-VL. UNITER uses 1.8M more image-caption pairs than Unicoder-VL, including in-domain dataset like Visual Genome Caption dataset during pre-training, which may greatly boost the performance of the image-text retrieval. Our Unicoder-VL can still achieve comparable results. 
\begin{table}[!htbp]
\centering
\resizebox{0.47\textwidth}{!}{
\begin{tabular}{ l  c  c  c  c  c  c }
 \toprule
\multirow{2}{*}{Methods} 
& \multicolumn{2}{c}{(Q$\rightarrow$ A)} 
& \multicolumn{2}{c}{(QA$\rightarrow$ R)} 
& \multicolumn{2}{c}{(Q$\rightarrow$ AR)} \\
& val & test & val & test & val & test \\
\midrule
R2C~\cite{zellers2019recognition} & 63.8 & 65.1 & 67.2 & 67.3 & 43.1 & 44.0 \\
\midrule
VisualBERT~\cite{li2019visualbert}$^\dag$ &70.8 &71.6 &73.2 &73.2 &52.2 & 52.4 \\
ViLBERT~\cite{lu2019vilbert}$^\dag$ & 72.4 &73.3 &74.5 &74.6 &54.0 &54.8 \\
B2T2~\cite{alberti2019fusion}$^\dag$ &71.9 &72.6 &76.0 &75.7 &54.9 &55.0 \\
VL-BERT~\cite{su2019vl}$^\dag$ &73.7 &74.0 &74.5 &74.8 &55.0 &55.5 \\
UNITER*~\cite{chen2019uniter}$^\dag$ &72.8 &- &75.3 &- &54.9 &- \\
\midrule
Unicoder-VL (w/o pre-training) &71.6 &- &73.1 &- &52.3 &-    \\
Unicoder-VL  &72.6 &73.4 &74.5 &74.4 &54.5 &54.9    \\
 \bottomrule \end{tabular}}
 \caption{Results compared to the state-of-the-art methods with single model on VCR dataset by the time of submission. $^\dag$ means concurrent works. * means that the UNITER's one-stage pre-training result, which is similar to the concurrent work's setting.}
\label{VCRresult}
\end{table}

\subsubsection{Results on Visual Commonsense Reasoning (VCR)}
Our final results on the VCR task are shown in
Tab~\ref{VCRresult}. Pre-training Unicoder-VL only slightly improves the performance. This might be because the pre-training task of image captioning is at the perceptual level, while the VCR task is at the cognitive understanding level. There is a gap between these two data types. Compared with baseline, R2C, we do not use task-specific modules. Instead, we simply add a simple classification layer to Unicoder-VL and jointly train the whole model end-to-end. Unicoder-VL outperforms R2C by large margins, indicating the power of our simple cross-modal architecture. 
The results without pre-training are slightly lower than results of pre-trained Unicoder-VL. It proves that VCR benefits from cross-modal pre-training. However, due to the difference of VCR dataset and caption dataset, the pre-training will not help too much.

Compared with other concurrent works, i.e. ViLBERT, VisualBERT, B2T2 and VLBERT,
our Unicoder-VL achieves comparable performance with state-of-the-art results. It proves that pre-train the tranformer-based model with large-scale dataset will yield improvement than previous task-specific methods on visual commonsense reasoning tasks. Note that UNITER proposes a two-stage pre-training for VCR. Here, we select the one-stage pre-training result of UNITER, and it shows similar performance with concurrent works. But two-stage pre-training may be helpful on some very different datasets, like VCR and the caption dataset.

\subsection{Discussion}  
For the pre-training tasks. Unlike VideoBERT \cite{sun2019videobert}, we do not use image-only inputs since the model fails to converge. But the viusal inputs of VideoBERT is actually generated visual words and its objective is still LM pre-training. We assume the true visual inputs without the guidance of linguistic data will damage the pretrained weights of BERT, which is pre-trained on linguistic data only. For future works, we are curious about how we could extend Unicoder-VL to image-only tasks like image-caption, scene graph generation or visual saliency detection.

For image-text retrieval task, the results of Unicoder-VL outperform all the methods without jointly pre-training (acturally viusal features from ResNet and linguistic word embeddings are pre-trained separately). It demonstrates that this transferring learning can also achieve great performance in cross-modal tasks. However, for image RoI based methods like SCAN\cite{lee2018stacked}, Unicoder-VL and ViLBERT \cite{lu2019vilbert}, the backbone of Faster-RCNN is still not fine-tuned with the whole model during cross-modal training. We have no idea that whether the performance is better or not if the backbone of detection model is fine-tuned with the cross-modal training and how to do so. We would like to explore these in the future.

We notice that the zero-shot image-text retrieval result of UNITER\cite{chen2019uniter} is much higher than ours. The reason is that UNITER uses in-domain dataset incluing MSCOCO Caption and Visual Genome Caption dataset to pretrain. These datasets are very similar to Flickr30k and it may be not a zero-shot testing. We believe that it is inappropriate to use in-domain dataset as pre-training dataset unless as the second-stage pre-training dataset because this in-domain dataset is human-annotated (of high quality) but Conceptual Captions and SBU Captions are automatically collected (sometimes not human-like or not related). However, we agree that the performance on these downstream tasks should be enhanced with more high-quality pre-training data. 

\subsection{Ablation Studies}
In this section, we perform ablation experiments in order to better understand the effect of the model size and the pre-train dataset size. 

\textbf{Effect of Model Size.}
We compare the results of Unicoder-VL models when varying Transformer encoder layers. We test our model with 6-layer, 12-layer and 24-layer Transformer encoders. If the number of the layers are less than 12, we simply load the first several layers of pre-trained weights from BERT. As shown in Tab~\ref{Transformerlayers}, we find that the image-text retrieval tasks benefit from larger models. 
\begin{table}[!htbp]
\centering
\resizebox{0.47\textwidth}{!}{
\begin{tabular}{ l  c  c  c  c  c  c }
 \toprule
\multirow{2}{*}{Methods} &

\multicolumn{3}{c}{Sentence Retrieval} & \multicolumn{3}{c}{Image Retrieval} \\
                  &  R@1 & R@5 & R@10 & R@1 & R@5 & R@10 \\
                    \midrule
    Unicoder-VL (6-layer)   &72.4 &	93.1 &96.3  &58.1 &	83.4 &90.2    \\
    Unicoder-VL (12-layer)   &86.2 &96.3 &99.0  &71.5 &90.9 & 94.9    \\
    Unicoder-VL (24-layer)   &86.5 &97.6 &99.3  &73.6 &92.3 & 95.8    \\
  \bottomrule \end{tabular}}
 \caption{Ablation study of the depth of Unicoder-VL with respect to the number of Transformer encoder layers. All of these experiments are fine-tuning on Flickr30k with pre-trained Unicoder-VL.}
\label{Transformerlayers}
\end{table}

\subsubsection{Effect of Training Sets Size} 
We also studied the impact of the size of the pretraining dataset. For this experiment, we take 75\% from the full dataset, and pretrain and finetune Unicoder-VL using the same setup as above. We can see that the accuracy grows monotonically as the amount of data increases, which suggests that Unicoder-VL may benefit from even more pretraining data. The same experiment results can be observed in ViLBERT \cite{lu2019vilbert} and UNITER \cite{chen2019uniter}.

\begin{table}[!htbp]
\centering
\resizebox{0.47\textwidth}{!}{
\begin{tabular}{ l  c  c  c  c  c  c }
 \toprule
\multirow{2}{*}{Methods} &

\multicolumn{3}{c}{Sentence Retrieval} & \multicolumn{3}{c}{Image Retrieval} \\
                  &  R@1 & R@5 & R@10 & R@1 & R@5 & R@10 \\
                    \midrule
    Unicoder-VL (0)   &73.0 &	89.0 &94.1  &57.8 &82.2 &88.9    \\
    Unicoder-VL (3M)   &82.3 &95.1 &97.8  &68.3 &90.3 & 94.6    \\
    Unicoder-VL (3.8M)   &86.2 &96.3 &99.0  &71.5 &90.9 & 94.9    \\
  \bottomrule \end{tabular}}
 \caption{Ablation study of the Flickr30k retrieval results of Unicoder-VL with respect to the pre-training dataset size. The number in parentheses 
 is the number of image-text pairs we used in pre-training. 0 means without pre-training.}
\label{Datasetsize}
\end{table}

\section{Conclusion}

\noindent In this work, we proposed Unicoder-VL for cross-modal tasks. We utilize large-scale image-caption pairs to pre-train Unicoder-VL. We introduce three different pre-training tasks to align the visual and linguistic modalities and learn better cross-modal representations. When fine-tuning on image and sentence retrieval tasks, our experiment results on Flickr30K and MSCOCO datasets demonstrate that our pre-trained Transformer model can boost retrieval performance significantly. The zero-shots experiments exhibit that Unicoder-VL can learn general cross-modal knowledge, which take effects in image retrieval and sentence retrieval tasks directly, without any task-specific fine-tuning. The VCR experiment shows that cross-modal pre-training improve the ability of visual commonsense reasoning. This pre-training method is general and not limited to these tasks. We do not see any reason preventing it from finding broader cross-modal applications, including video related tasks. Meanwhile, we still have interest on how Unicoder-VL learn from image-only inputs. We will try to extend to some image-only tasks like image-caption and scene graph generation in the future work.

\section{Acknowledgments}
We thank the anonymous reviewers for their helpful comments and discussions. This research is supported by National Natural Science Foundation of China under Grant NO.61672062, NO.61232005.

\bigskip

\bibliographystyle{aaai}
\bibliography{AAAI}
\end{document}